\documentclass{article}

\usepackage[preprint, nonatbib]{neurips_2020}

\usepackage[utf8]{inputenc} 
\usepackage[T1]{fontenc}    
\usepackage{hyperref}       
\usepackage{url}            
\usepackage{booktabs}       
\usepackage{amsfonts}       
\usepackage{nicefrac}       
\usepackage{microtype}      
\usepackage{subcaption}
\usepackage{wrapfig}

\usepackage{graphicx}
\usepackage{amsmath}

\newcommand{\eq}{\begin{equation}}
\newcommand{\eqx}{\end{equation}}
\newcommand{\eqn}{\begin{eqnarray}}
\newcommand{\eqnx}{\end{eqnarray}}

\newcommand{\qq}{\quad\quad}

\newcommand{\gchar}[1]{{\tt #1}}
\newcommand{\DD}{{\mathcal D}}
\newcommand{\GG}{{\mathcal G}}

\title{Neural networks adapting to datasets:\\ learning network size and topology}

\author{Romuald A. Janik\\
Institute of Theoretical Physics\\
Jagiellonian University\\
\L{}ojasiewicza 11, 30-348 Krak\'{o}w, Poland.\\
\texttt{romuald.janik@gmail.com}\\
\And
Aleksandra Nowak\\
Faculty of Mathematics and Computer Science\\
Jagiellonian University\\
\L{}ojasiewicza 6, 30-348 Krak\'{o}w, Poland\\
\texttt{nowak.aleksandrairena@gmail.com}}

\begin{document}

\maketitle

\begin{abstract}
We introduce a flexible setup allowing for a neural network to learn both its size and topology during the course of a standard gradient-based training. The resulting network has the structure of a graph tailored to the particular learning task and dataset. 
The obtained networks can also be trained from scratch and achieve virtually identical performance. We explore the properties of the network architectures for a number of datasets of varying difficulty observing systematic regularities. The obtained graphs can be therefore understood as encoding nontrivial characteristics of the particular classification tasks.
\end{abstract}

\section{Introduction}

Classical Artificial Neural Networks typically have a fixed and rather rigid structure \emph{a-priori} independent of the specific learning task.
In addition, a network usually utilizes all of its resources irrespective of whether the given learning task is challenging or not. One might expect, however, that the \emph{natural} network structure and size for a specific dataset should depend on the characteristics of the corresponding classification task.  

The goal of this paper is to propose a simple and flexible framework in which, during the course of standard gradient-based training, the network learns not only the classification task, but also its effective structure and size.
In order to do that, we would like to follow a particular path, which allows to successively lift the structural rigidity assumptions of classical neural networks, while retaining significant flexibility.

In order to implement very general architectures of information processing, it is convenient to be able to base a neural network architecture on an \emph{arbitrary} (directed acyclic) graph. This has been first done for the distinct resolution stages of an ImageNet network in \cite{xie2019exploring}, while the work~\cite{janik2020neural} extended the construction to the whole network from the input to the final output layer. 

Basing the architecture on an arbitrary graph has the benefit of allowing complete flexibility in prescribing both local (short-range) and global (long-range) information processing pathways, while keeping the elementary computational node relatively simple and fixed. This is in contrast to many Neural Architecture Search (NAS) investigations which, on the other hand, concentrate on exploring the intricate structure of a computational cell while not necessarily allowing full flexibility of the global structure (see e.g. \cite{ying2019bench, zhang2018you}).

The emphasis of~\cite{janik2020neural} was on investigating how the performance of a wide variety of neural network architectures based on various random graphs depends on the characteristics of the graphs themselves. 
A family of well performing architectures was identified (the so-called \emph{quasi-1-dimensional networks}), although other examples of good networks were also observed (among them notably some based on fMRI networks). 

In this paper, we would like to take the next step and allow the network to choose its own size and topology during training. Of course, the details of the answer depend on the precise way that topology change is incorporated in the learning process. In our setup, the topology is encoded in the magnitudes of the weights associated with the edges of the graph. In consequence, 
we can adopt here the simplest possible and most standard choice of adding a sparsity inducing L1 loss.
Implementing this loss for the graph edges has a very natural interpretation of enforcing that the network architecture is most economical i.e. has the smallest number of active connections. 
Therefore during training one seeks a trade-off between performance and utilized resources.
The above framework also naturally incorporates the possibility of reducing the network's effective size, as turning off edges can also lead to erasing parts of the network containing several computational nodes.

Another, quite different motivation for exploring this question is the inspiration from neuroscience, where it is claimed that neuronal connections get pruned during sleep~\cite{tononi2013perchance}. The sparsity induced loss mimicks, in a \emph{very qualitative} way, this phenomenon. One may also expect that biological neuronal networks seek to minimize the overall energy cost of computation.

The emergence of a particular network structure and size
is especially interesting in view of training the network on various datasets. Intuitively we might expect that more difficult classification tasks would require larger, more complicated networks. The present setup allows us to investigate exactly this issue. 

Apart from practical applications, the construction of the present paper has an additional theoretical benefit of quantifying, in a nontrivial way, the internal structure of a dataset or, more generally, learning task (see~\cite{complexity} for a similar perspective).

\section{Related work}

\paragraph{Network pruning.}

The issue of introducing sparsity into the neural network has been extensively explored in the field of neural network pruning, which aims at removing redundant parameters from the model in an effort to produce compressed and resource efficient network without a significant loss in performance \cite{lecun1990optimal, han2015learning}.  Multiple approaches pursuing this goal have been proposed over the recent years \cite{lecun1990optimal,  han2015learning, hassibi1993second, wen2016learning, molchanov2016pruning, narang2017exploring}, either specially designed to eliminate whole structures \cite{wen2016learning,  molchanov2016pruning, narang2017exploring, anwar2017structured, li2016pruning}, or uniformly erasing all parameters of the network satisfying a certain condition, regardless of their role in the model \cite{lecun1990optimal, han2015learning, hassibi1993second, lin2017runtime} -- see \cite{cheng2017survey, blalock2020state} for a survey. Most of the proposed pruning techniques  incorporate an adjusting procedure in which the learned sparse model is fine-tuned in an attempt to match the performance of the uncompressed model \cite{han2015learning, wen2016learning, molchanov2016pruning,  li2016pruning}. At the same time, the pruned architectures have been repeatedly reported to be unable to reproduce their results when trained from scratch with a different weight initialization \cite{han2015learning, li2016pruning}. This belief has been recently at the centre of some debate. In particular, \cite{frankle2018lottery} hypothesize that pruning is merely a way of retrieving an optimal subnetwork for a given input weight initialization, while \cite{liu2018rethinking} argue that perhaps the structurally-pruned architecture itself is most important for the efficiency of the final model and can achieve high results even with a weight reinitialization. Moreover, different pruning methods tend to behave inconsistently depending on the difficulty of the task \cite{liu2018rethinking, gale2019state}. These results motivate research into a better understanding of the connection between the original network and the learned sparse architecture with respect to a given task and weight reinitialization,  indicating a link between neural architecture search and pruning approaches.  

\paragraph{Learning the network's wiring topology.}

Neural architecture search (NAS) aims at automatically designing the network topology and block operations that allow to achieve high performance \cite{zoph2016neural, baker2016designing, real2019regularizedAmoebaNet, elsken2019nas_survey}. An interesting approach to this problem has been proposed by \cite{xie2019exploring}, who base different stages of computations in the network on the topology of arbitrary, random  directed acyclic graphs. This has been further exploited by \cite{janik2020neural}, who additionally allow for arbitrary interstage connections and examine the correlations between network performance and graph characteristics of the resulting architectures.
In both of these papers, however, the network's architecture was not learned.
Graph architectures has been also studied by \cite{wortsman2019discovering}, who directly learn the networks wiring topology by adding and erasing graph edges during the training. Finally, in \cite{zhang2018you} the authors propose a pruning technique which reduces the fully connected cell operation block to a sparse counterpart. Such cell blocks are then stacked together in a sequential manner to form the resulting architecture, as is often done in typical NAS algorithms \cite{ying2019bench, real2019regularizedAmoebaNet}. 

\paragraph{The contribution of the present paper.}

In this paper we implement a procedure which allows the network to learn both its \emph{size} and \emph{connectivity} for a given classification task during the course of training and we analyze the structure of the obtained networks as a function of the dataset.

We encourage the network to find sparse connectivity for a given task by imposing L1-regularization on the edges of the fully connected DAG network defined using the framework proposed by \cite{janik2020neural}. Note that by modelling the \emph{whole} network as a fully connected DAG (instead of only the cells, as in \cite{zhang2018you} or distinct resolution stages as in~\cite{xie2019exploring}), we allow for the occurrence of complex connectivity patterns between different stages of computation (as well as within each stage). Consequently, the appearance of any bottlenecks, sequentiality or other structures in the information flow should emerge naturally from the training procedure and not as an explicit bias introduced by the researcher. Also the size and ultimate sparsity of the obtained network is not set from the outset (e.g. as the parameter $k$ in ~\cite{wortsman2019discovering}), but arises in the process of learning and varies with the complexity of the given
classification task.

We concentrate on the following questions:
\begin{enumerate}
\item What are the common features of the resulting networks?
\item How do the properties of the network depend on the specific dataset and its difficulty?
\item Do the obtained architectures perform well when trained again from scratch with an arbitrary initialisation?
\end{enumerate}

A positive answer to the third point would show that the learned networks are indeed meaningful for the dataset in question and thus making the answers to the first two questions more relevant.  Additionally, this would support the suggestion that structural pruning can indeed lead to the uncovering of optimal architectures \cite{liu2018rethinking}.

\section{The setup}
\label{s.setup}

\paragraph{From a DAG to a neural network.}

\begin{figure}
\centering
\subcaptionbox{\label{fig:node}}%
  [.45\linewidth]{\includegraphics[height=2.5cm]{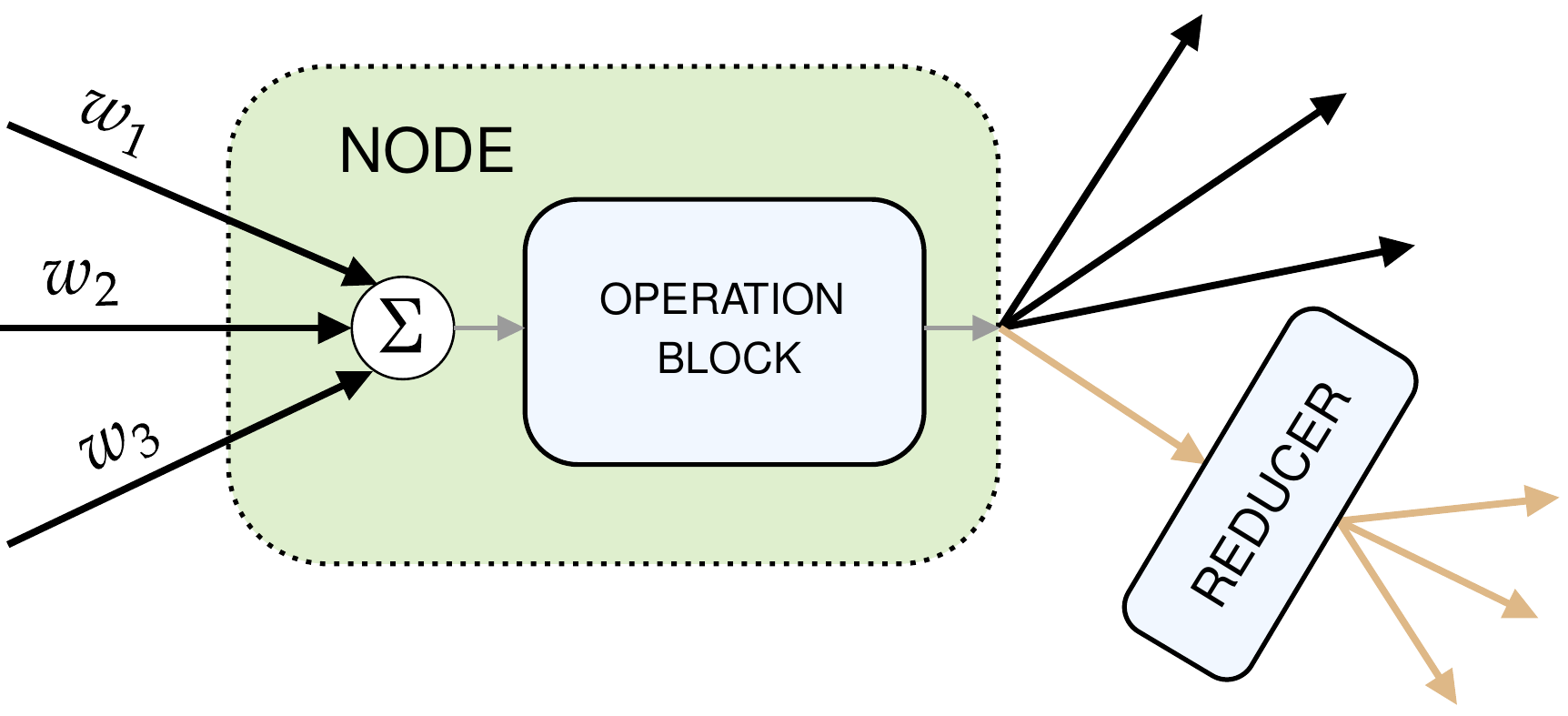}}
 \subcaptionbox{\label{fig:fulldag}}
  [.25\linewidth]{\includegraphics[height=3cm]{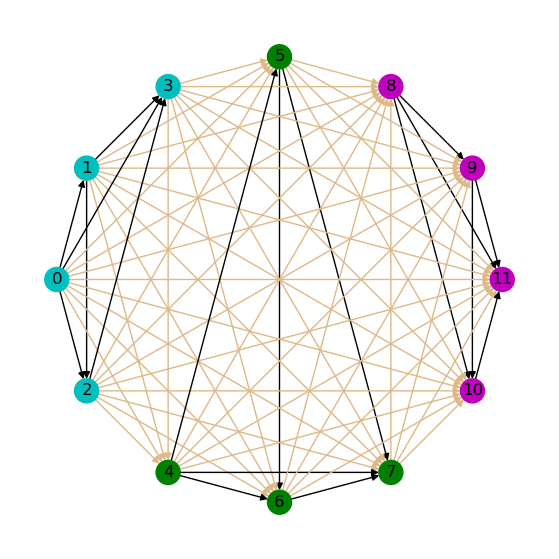}}
\subcaptionbox{\label{fig:gnaw}}
  [.25\linewidth]{\includegraphics[height=3cm]{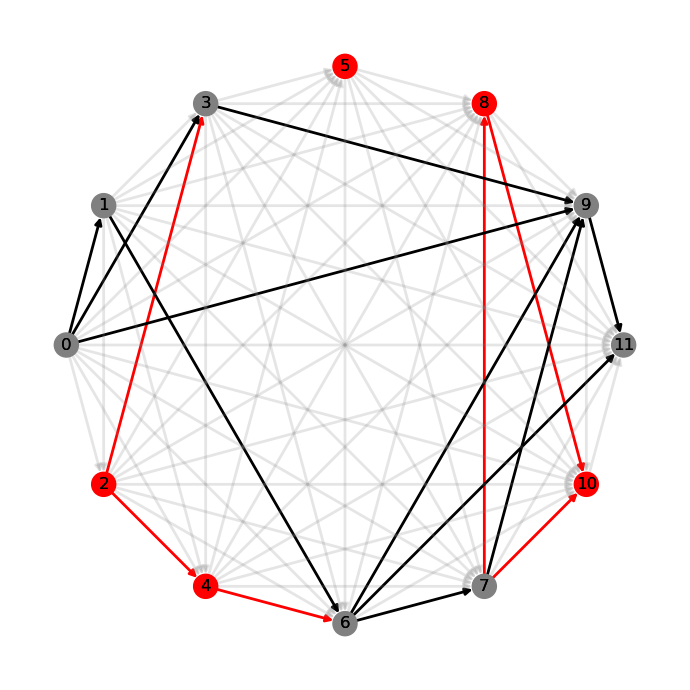}}
  \caption{\textbf{(a)} The architecture of a single node in the graph. The black arrows represent connections within a single resolution stage, while the beige ones go to lower resolution stages. \textbf{(b)} The architecture of a fully connected DAG (for clarity we use a $12$-node graph as an example). Different node colours represent different stages of computations. \textbf{(c)} An example of the thresholding procedure on the DAG from Fig. \ref{fig:fulldag}. The light gray connections are the edges erased from the graph by thresholding. The red edges and nodes are the paths that do not contribute to the output and can be further removed. } \label{fig:1}
\end{figure}

We construct the neural network architecture using a fully-connected, directed acyclic graph with 60 nodes, following the approach in \cite{janik2020neural}. The nodes correspond to a stack of operations performed on tensors, while the edges define how the information is propagated in the network (see Fig.~\ref{fig:node}). Each node performs a block of transformations composed of a weighted aggregated sum of the inputs, followed by a ReLU nonlinearity, Conv-2D mapping and a BatchNorm layer.  A residual connection in the form of Conv-1x1 connects the output of the weighted aggregated sum and the output of the whole block. The weights in the aggregation performed in the nodes are given by the weights defined on the edges of the DAG and are trainable as well. Intuitively, each such edge weight corresponds to the portion of the outgoing node’s output that is incorporated in the ingoing node’s computations. 

The whole network is divided into three, equally-sized stages. Within each stage, all nodes perform computations on tensors with the same spatial resolution and number of channels. The first (input) node sets the initial number of channels $C$ for the first stage, keeping the same spatial resolution as the size of the input image. The second and third stages downsample the resolution and increase the number of channels by a factor of 2 and 4, respectively.  This transition is performed on every edge that connects different stages with the use of a reduce block as defined in \cite{janik2020neural}  -- see Fig.~\ref{fig:node} and Fig.~\ref{fig:fulldag}. Finally, when the computation reaches the last node in the network, a standard global average pooling is applied, followed by a linear transformation to the desired output dimension. 

\paragraph{Imposing sparsity.}

As described above, the connectivity of the neural network is encoded in the weights associated to the edges of the directed acyclic graph. A change in the network topology would arise through a number of those edge weights being set to zero as a consequence of the learning process. An obvious way to enforce that is to impose sparsity by a L1 loss on the edge weights.

In the work \cite{janik2020neural}, following \cite{xie2019exploring}, the final weights of the edges were obtained by evaluating a \emph{sigmoid} on the underlying network weights, which ensured that the effective edge weight was in the range $[0,1]$. It turns out, however, that the $sigmoid$ prevents the L1 loss from working as intended. 
Therefore, in the present work, we use a \emph{tanh} activation instead of 
the \emph{sigmoid} and define the sparsity loss as
\eq
L_{sparsity}(w) = \sum_{e \in edges} |\tanh w_{e}|
\eqx
Since $\tanh$ is linear in the vicinity of zero, the problem with the sparsity loss is cured.
Clearly, the sign of the final weight does not matter, thus we will always analyze $|\tanh w_e|$ below.

As we would like the final network architecture to depend purely on the \emph{learning process}, we choose a constant initialisation of all edge weights so that $\tanh w_e^{init} = 0.5$. In this way we do not bias the edges which could have been set by the random initialisation to values close to zero and in consequence be easier to be turned off by the learning procedure.  

\paragraph{Pruning the network.}

As a result of adding $L_{sparsity}$ to the objective loss, the edge weights are encouraged to have small magnitude.  Edges with very low magnitude may be interpreted as paths that do not contribute much to the computation of the network and are therefore redundant and can be safely eliminated.

In order to isolate the active subgraph within the fully connected DAG, we perform straightforward thresholding at the end of the training. Each edge with the absolute value of the weight smaller than a given threshold $\tau$ is erased from the graph. Note that such a procedure may lead to paths that do not originate in the input node or do not end in the output node (see Fig.~\ref{fig:gnaw}). Those "dead paths" can be further removed from the graph, as they do not contribute to the network output. This can be implemented by recursively removing zero input-degree and zero output-degree nodes that are not the first or last node in the graph, which is a simple variation of the Kahn's algorithm for topological sort~\cite{kahn1962topological}. The thresholding procedure may also completely disconnect the graph -- in such a case the networks performance drops to match the one of a random predictor. In order for the chosen threshold to be acceptable, we require the performance of the thresholded network to be comparable to the original fully connected DAG network.

\section{Results}

\paragraph{Datasets utilised in the current work.}

One of the main goals of the present paper is to investigate how the learned architectures depend on the difficulty of the classification task. This necessitates the use of a range of datasets of varying complexity, while keeping the architecture of the initial network unchanged.

Since the employed neural network architectures were designed for operating on CIFAR10 data ($32 \times 32$ colour images, belonging to 10 classes), we use a set of datasets from~\cite{complexity}, which transformed the MNIST, KMNIST and FashionMNIST datasets into the desired format (denoted below with a prefix 'C'). This was achieved by randomly embedding the input image into the $32 \times 32$ resolution format, randomly colouring it, and adding some colour noise. These datasets are significantly easier than CIFAR10. In addition, we also use a teared up version of CIFAR10 from~\cite{complexity} (denoted as CIFAR10\_T8), where the original images were cut into $8\times 8$ pieces and then shuffled and rotated by multiples of $90^\circ$. The random shuffling and rotations were kept \emph{fixed for all} images in CIFAR10. This dataset is more challenging than the original CIFAR10. Thus the progression of difficulty of the employed datasets is as follows:

\eq
\text{CMNIST} < \text{CKMNIST} < \text{CFashionMNIST} < \text{CIFAR10} < \text{CIFAR10\_T8}
\eqx

We use the same learning regime for all datasets, which is identical to the one used by~\cite{janik2020neural} for CIFAR10 (see \textit{Supplementary Materials C}). We train the fully connected DAG network with 10 different random initialisations for each dataset. The initial number of channels is set to $C=11$ in order for the networks to have approximately the same number of parameters as ResNet-56~\cite{he2016deep_residual}. The models are trained and evaluated on the standard train/test splits defined for each dataset.\footnote{for CIFAR10\_T8 we adopt the same split as for CIFAR10.} The objective loss function is composed of the standard cross-entropy loss to which we add the  $L_{sparsity}$ loss term with coefficient $\lambda=1e-3$.

\paragraph{Thresholding and test accuracies.}

\begin{figure}[ht]
   \centering
    \subcaptionbox{\label{fig:sparsity}}%
   [.50\linewidth]{\includegraphics[height=2.5cm]{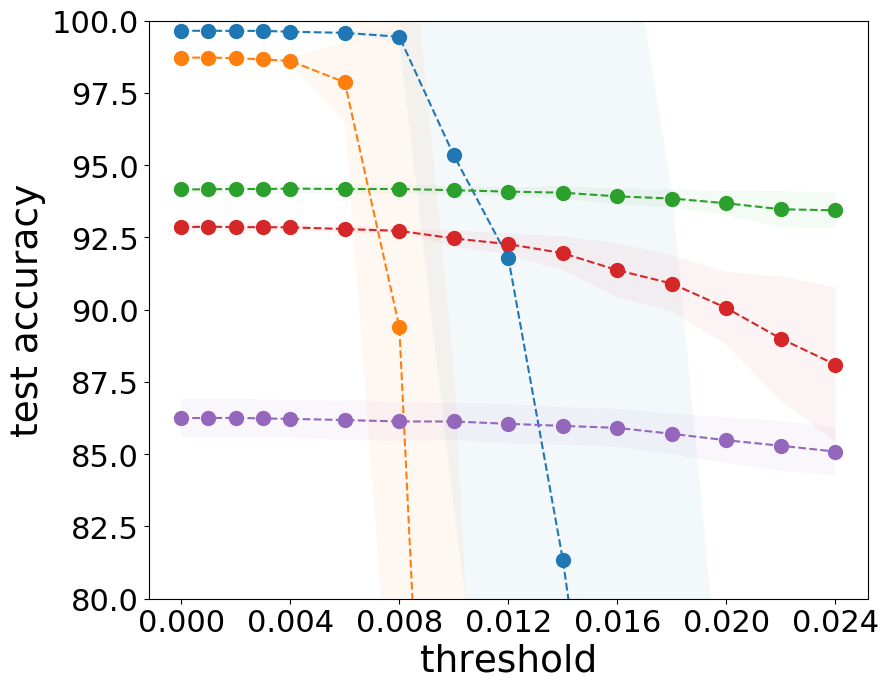}
    \includegraphics[height=2.5cm]{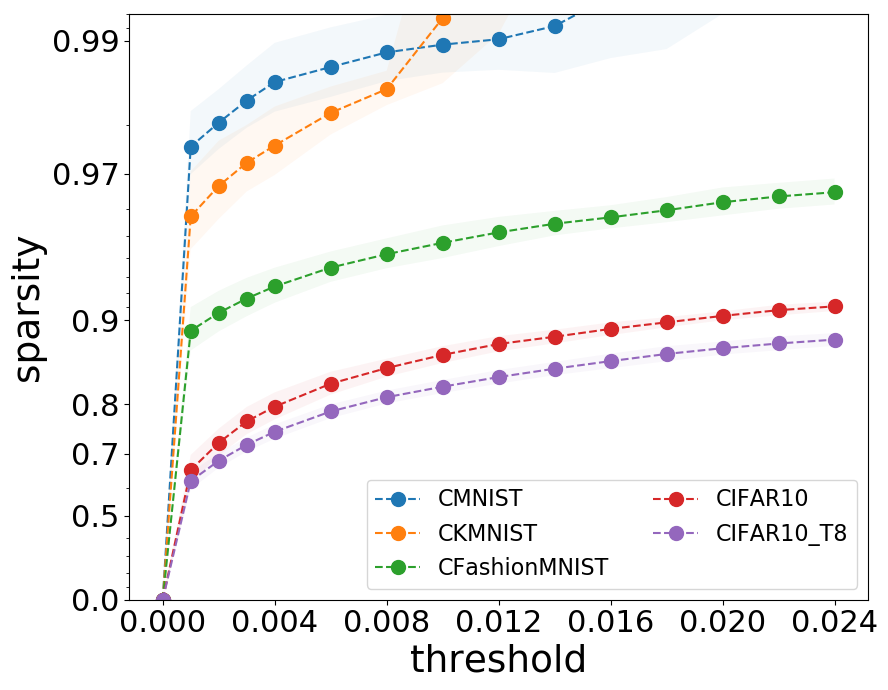}}
   \subcaptionbox{\label{fig:retrain}}%
   [.48\linewidth]{\includegraphics[height=2.5cm]{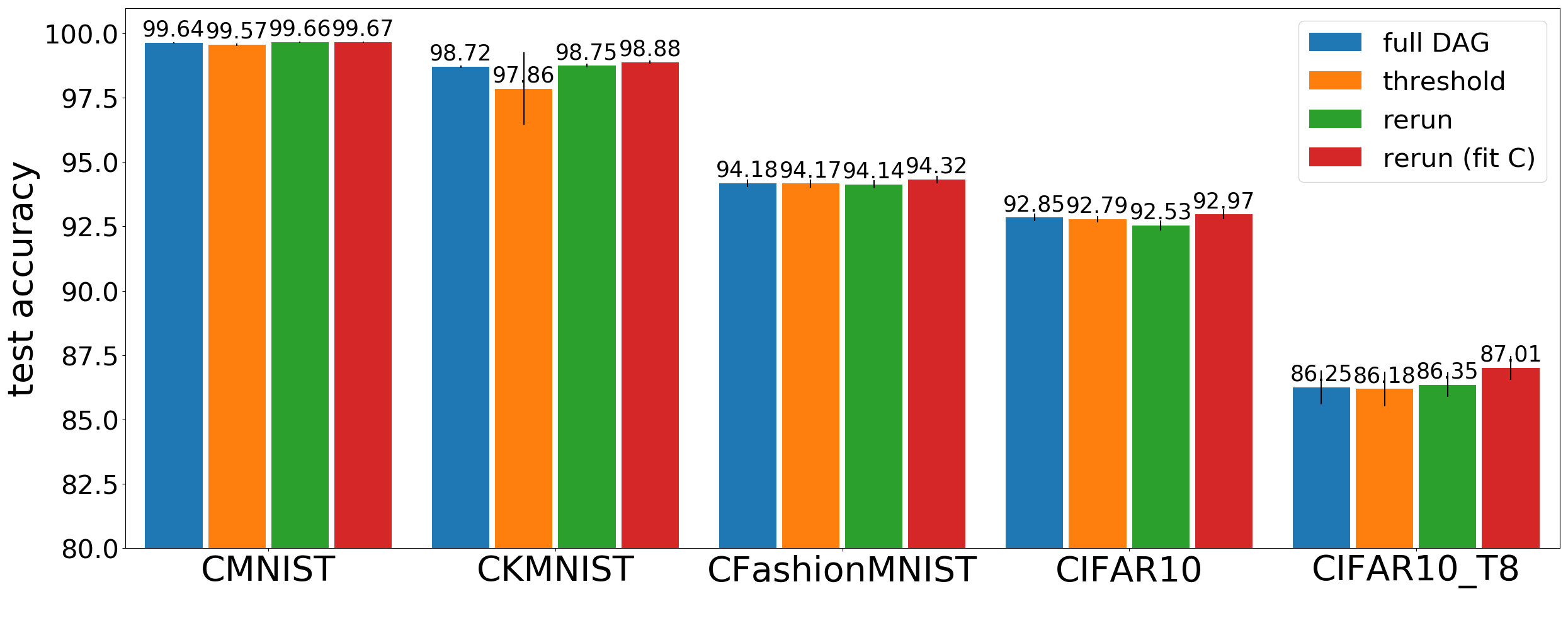}}
   \caption{\textbf{(a)} The test accuracy and sparsity plots for given thresholds. \textbf{(b)} The test accuracy of the initial fully-connected DAG, the thresholded model,
   the thresholded model after retraining from scratch, and same but with $C$ increased to match the number of parameters of ResNet-56.}

  \end{figure}

For a set of predefined thresholds $\tau$ we compute the mean test accuracy and mean sparsity obtained after downsizing the graph as described in section~\ref{s.setup}. The results are presented in Fig. \ref{fig:sparsity}. For relatively simple datasets such as CMNIST and CKMNIST, the thresholding procedure quickly increases sparsity (note that for CMNIST and threshold $\tau=0.008$ both the sparsity and test accuracy are approximately 99\%). Further increase in the sparsity causes the model to deteriorate and finally disconnects the graph, as pictured on the plots. More challenging datasets, as expected, seem to require higher connectivity. For all the datasets erasing edges up to threshold $0.006$ practically does not affect the accuracy, while significantly improving the sparsity (note the logarithmic scale on the sparsity plot). This suggests that the deleted links are indeed insignificant and redundant.

For the rest of the paper, we investigate the architectures obtained with the threshold $0.006$ as a reasonable choice for all datasets, which allows for the maximum sparsity without a noticeable drop in performance. We should emphasise that we do \emph{not} perform (and do not need) any further fine-tuning in contrast to many pruning approaches. The architectures obtained by us are also capable of achieving high performance when trained again from scratch with arbitrary initialisation. We investigate this situation in the following paragraph.

\begin{wrapfigure}[10]{r}{0.5\textwidth}
\vspace{-0.4cm}
\hfill\includegraphics[height=2.5cm]{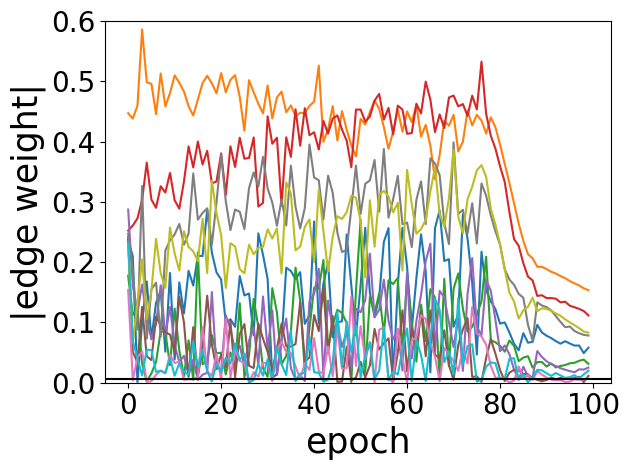}
\hfill\includegraphics[height=2.5cm]{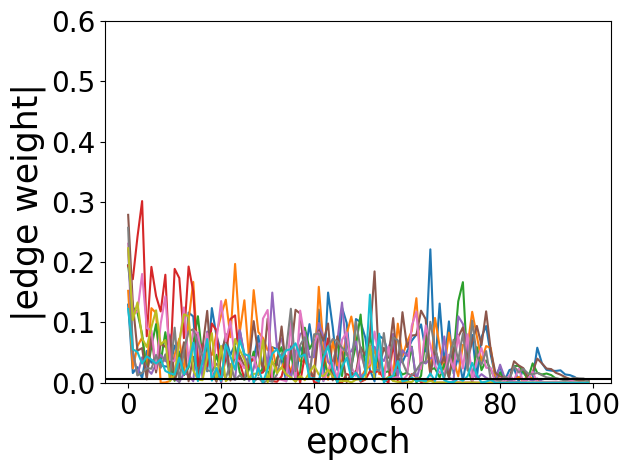}
\hfill\mbox{}
  \caption{The magnitude of the retained (left) and eliminated (right) edges during the training.}
  \label{fig.weight_evolution}
\end{wrapfigure}

We consider two scenarios. 
In the first one we train the thresholded network from scratch using different initialisation. In the second setup we additionally fit the number of initial channels $C$ separately for each of the thresholded networks, so that the number of parameters approximately matches the number of parameters of ResNet-56 (and the original, fully connected DAG network).
As observed in Fig. \ref{fig:retrain}, retraining the obtained architectures with the same learning regime yields comparable or better performance to the one achieved by the fully connected DAG. This shows that the resulting connectivity patterns are indeed meaningful and not limited to a specific weight initialisation. 
Moreover, in both retraining scenarios, the time of one epoch is significantly smaller than for the complete DAG - see \textit{Supplementary Materials A}.

\paragraph{Edge weight dynamics.}

Before we analyse in detail the characteristics of the thresholded networks for various datasets, we would like to comment on the behaviour of the magnitudes of the edge weights during training. In Fig.~\ref{fig.weight_evolution}, we show the evolution of a subset of weights which get retained after thresholding and of those which get eliminated.
The weights fluctuate, which can be interpreted as exploring various wiring patterns,
and eventually settle down to the final values (thus picking out a particular connectivity pattern) in the last stage of training after the reduction of the learning rate.

\paragraph{Networks obtained for various datasets.}

\begin{figure}
  \centering
\begin{tabular}{ccccc}
\includegraphics[height=3.5cm]{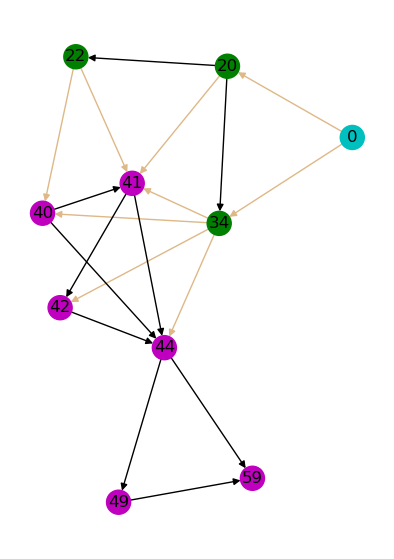} &
\includegraphics[height=3.5cm]{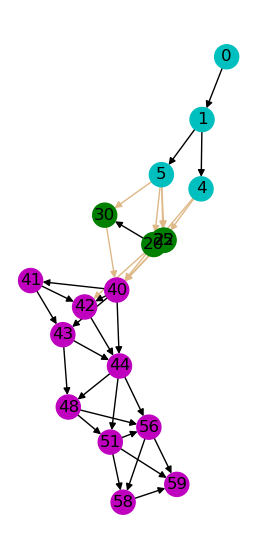} &
\includegraphics[height=3.5cm]{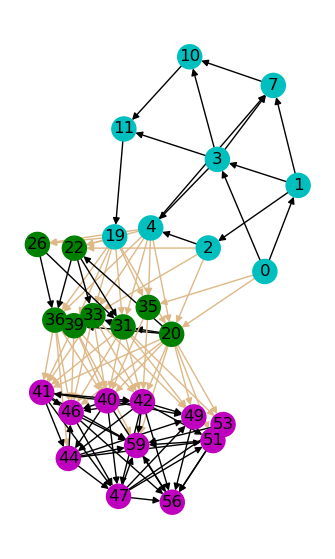} &
\includegraphics[height=3.5cm]{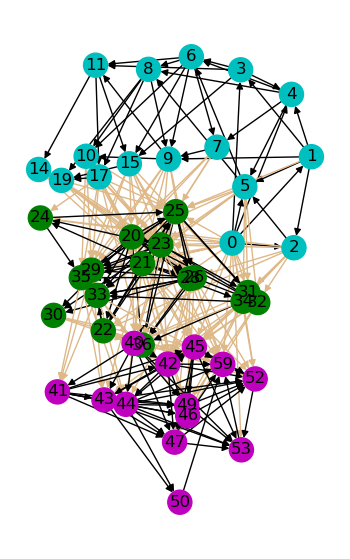} &
\includegraphics[height=3.5cm]{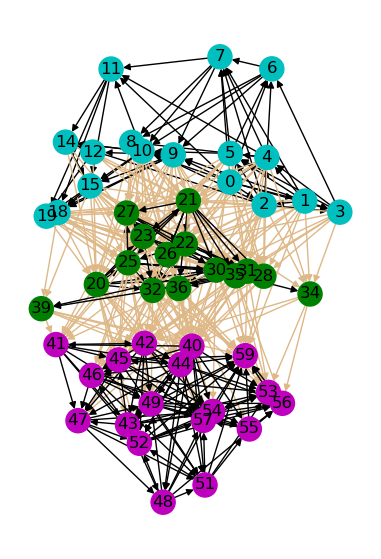} \\
CMNIST &
CKMNIST &
CFashionMNIST &
CIFAR10 &
CIFAR10\_T8
\end{tabular}
  \caption{The thresholded networks (at the threshold 0.006) obtained from training for a specific random initialisation. See \textit{Supplementary Materials B} for other initialisations.}
  \label{fig.networks}
\end{figure}

In Fig.~\ref{fig.networks}, we show a selection of networks obtained for the various datasets for specific random initialisation of the standard network parameters. Recall, however, that the initial values of all edge weights were set to $0.5$ in order not to bias the emerging network connectivity by random initialisation.
We observe visually a steady increase in the network complexity with the difficulty of the classification task. In addition, we see significant structural similarity between the graphs obtained for the same dataset from different random initialisations (see Fig. 7. in the {\it Supplementary Materials B}).
In the following we will provide more quantitative investigations into these issues.

\begin{table}[t]
  \caption{Main structural characteristics of the obtained networks (with threshold 0.006) averaged over 10 different initializations. The initial network has in total 60 nodes, distributed evenly between the three stages and has 863.3k parameters.}
  \label{tab.structural}
  \centering
  \footnotesize
\begin{tabular}{lrrrrr}
\toprule
 &     CMNIST &  CKMNIST &  CFashionMNIST &  CIFAR10 &  CIFAR10\_T8 \\
\midrule
sparsity (all) & $0.988 \pm 0.003$ & $0.982 \pm 0.004$ &  $0.935 \pm 0.008$ & $0.831 \pm 0.016$ &  $0.788 \pm 0.012$ \\
- stage 0      & $0.981 \pm 0.015$ & $0.983 \pm 0.009$ &  $0.905 \pm 0.023$ & $0.791 \pm 0.037$ &  $0.712 \pm 0.034$ \\
- stage 1      & $0.989 \pm 0.006$ & $0.989 \pm 0.006$ &  $0.936 \pm 0.028$ & $0.725 \pm 0.058$ &  $0.754 \pm 0.057$ \\
- stage 2      & $0.952 \pm 0.020$ & $0.905 \pm 0.012$ &  $0.822 \pm 0.035$ & $0.678 \pm 0.046$ &  $0.573 \pm 0.061$ \\
nodes (all)       & $12.5 \pm 2.3$ & $17.0 \pm 2.1$ & $26.9 \pm 2.3$ & $48.5 \pm 2.6$ & $48.2 \pm 1.5$ \\
- stage 0    & $3.6 \pm 1.8$ & $3.5 \pm 1.2$ & $8.8 \pm 1.0$ & $15.6 \pm 0.7$ &  $16.5 \pm 1.1$ \\
- stage 1    & $2.9 \pm 1.1$ & $3.7 \pm 1.3$ & $7.4 \pm 1.5$ & $17.3 \pm 1.6$ &  $14.9 \pm 1.3$ \\
- stage 2    & $6.0 \pm 1.3$ & $9.8 \pm 0.9$ & $10.7 \pm 1.5$ & $15.6 \pm 1.8$ & $16.8 \pm 1.2$ \\
parameters & $166k \pm 30k$ & $253k \pm 34k$ & $333k \pm 34k$ & $615k \pm 49k$ & $613k \pm 24k$ \\
\bottomrule
\end{tabular}\end{table}

\paragraph{Overall structural characteristics.} In Table~\ref{tab.structural} we have collected the means and standard deviations of the main characteristics of the obtained thresholded networks (with threshold 0.006). These data quantify several important tendencies which can be visually assessed from Fig.~\ref{fig.networks}.

Firstly, as the learning task gets easier the networks drop a larger fraction of the original connections, becoming much simpler. Similarly, the size of the network, as measured by the number of surviving nodes, increases with the complexity of the dataset. In this respect, the comparison of CIFAR10 and CIFAR10\_T8 is particularly interesting. Although the average size of the networks for these two datasets is virtually identical, the networks for the more challenging dataset are more interconnected as measured by the smaller sparsity fraction. Even then, however, the networks are much more sparsely connected than the initial fully connected DAG network, retaining at most around one fifth of the total number of connections.
It is also illuminating to look at the number of erased connections \emph{within} each stage separately. We see that the smallest sparsity is systematically in the last processing stage, thus the high level connections are most valuable for the network.

Secondly, the distribution of surviving computational nodes between the various resolution stages of processing differs systematically between the datasets. Of particular note is the relatively large number of nodes for KMNIST (Kuzushiji-MNIST) in the final stage of processing.  In a way this is natural, as the Japanese characters seem to require integration of features at the global level. An opposite tendency is represented by CFashionMNIST, where, on the other hand, the whole increase in the number of nodes with respect to KMNIST occurs essentially in the lower resolution stages. 

It is also interesting to compare from this perspective CIFAR10 and CIFAR10\_T8. They differ most in the number of nodes in the middle stage, with the latter dataset having the smaller number. The operation of tearing up preserves the features at smallest scales, significantly upsets the structure at mid resolutions where the boundaries of various patches would appear, and is more challenging at the global level in order to integrate the information from the scattered patches into the classification result. It is therefore tempting to speculate, that the features in the middle stage may be least useful and hence the network prefers more connections in the initial and final stage.

An interesting common feature of the networks for these two most challenging datasets is that even though the number of nodes of the ``low-level'' stage~0 and ``high-level'' stage~2 is essentially identical, there is a significant difference in the number of retained connections within these stages. The high-level stages have lower sparsity so are more interlinked and thus more complex.

Finally, it should be emphasised that the obtained networks are significantly smaller and contain a smaller number of parameters than the original fully connected DAG (whose number of parameters was fixed to be approximately equal to the one of ResNet-56). This is especially evident for the simple classification tasks, but even the networks for the most challenging datasets retain only $70\%$ parameters of the initial network, yet they do not suffer from a performance loss.

\begin{table}
  \caption{A selection of graph characteristics of the networks thresholded at 0.006.}
  \label{tab.graphchar}
  \centering
  \footnotesize
\begin{tabular}{lrrrrr}
\toprule
dataset &  CMNIST &  CKMNIST &  CFashionMNIST &  CIFAR10 &  CIFAR10\_T8 \\
\midrule
log\_paths         &  $ 4.03 \pm 1.03$ & $ 5.50 \pm 0.87$ & $ 11.76 \pm 0.81$ & $ 18.47 \pm 1.31$ & $ 21.37 \pm 1.27 $\\
mean\_path         &  $ 7.44 \pm 1.05$ & $ 8.95 \pm 0.76$ & $ 11.28 \pm 0.85$ & $ 16.42 \pm 0.99$ & $ 18.21 \pm 1.25$ \\
max\_path          &  $ 9.60 \pm 1.35$ & $ 12.00 \pm 1.41$ & $ 18.80 \pm 1.75$ & $ 28.00 \pm 2.49$ & $ 32.60 \pm 2.63$\\
ln\_communicability & $-3.26 \pm 1.86$ & $-3.86 \pm 1.28$ & $ 5.52 \pm 1.37$ & $ 11.03 \pm 0.90$ & $ 14.59 \pm 0.74$\\
edge\_connectivity &  $ 1.30 \pm 0.48$ & $ 1.30 \pm 0.48$ & $ 4.80 \pm 2.49$ & $ 13.60 \pm 3.44$ & $ 20.20 \pm 1.99$ \\
mean\_degree       &  $ 3.48 \pm 0.46$ & $ 3.75 \pm 0.30$ & $ 8.53 \pm 0.67$ & $ 12.31 \pm 0.86$ & $ 15.58 \pm 0.73$\\
pca\_elongation    &  $ 0.78 \pm 0.15$ & $ 0.72 \pm 0.10$ & $ 0.52 \pm 0.06$ & $ 0.37 \pm 0.06$ & $ 0.44 \pm 0.06$\\
\bottomrule
\end{tabular}
\end{table}

\paragraph{Graph characteristics.} In Table~\ref{tab.graphchar}, we have collected a selection of numerical graph characteristics which differ the most among the obtained networks. \gchar{log\_paths} is the logarithm of the total number of different paths going between the input node and output node. We observe a steady increase with the complexity of the dataset, which is in fact very natural. The same tendency occurs for other path related quantities like the mean length of the path, the maximal path between the input and output and \gchar{ln\_communicability}. \gchar{ln\_communicability} is (the logarithm of) the communicability -- an element of the exponent of the adjacency matrix which measures a weighted sum over walks between the input and output node \cite{estrada2008communicability}.

\gchar{edge\_connectivity} is the minimal number of edges which have to be removed so that the graph splits into disconnected components. It can be understood to measure the breadth of the graph, or the extent of interconnections and its increase for the more challenging datasets is clear. Similarly the mean degree of nodes also rises systematically.

The last observable in Table~\ref{tab.graphchar}, \gchar{pca\_elongation} was introduced in~\cite{janik2020neural}.
In that paper, a large ensemble of around 1000 networks with architectures based on various graphs was studied. One of the main results of~\cite{janik2020neural}, was the identification of a class of well-performing networks with a \emph{quasi-1-dimensional} structure. Since none of the classical graph features by itself could pick out good networks, \cite{janik2020neural} introduced a new graph characteristic, \gchar{pca\_elongation}, which numerically characterised the \emph{quasi-1-dimensional} structure.
\gchar{pca\_elongation} is defined in terms of the Kamada-Kawai graph embedding~\cite{kamada1989algorithm}, which serves to minimise the total energy associated to the lengths of the edges. Quasi-1-dimensional graphs have an elongated embedding which can be quantified by performing PCA on the set of node coordinates and looking at the explained variance ratio of the first component. The explicit definition is $\gchar{pca\_elongation} = 2\times variance\_ratio[0] - 1$.

The criterion for \emph{quasi-1-dimensional} graphs (denoted as Q1D) given in~\cite{janik2020neural} 
is

\eq
\gchar{pca\_elongation} > 0.25 \qq \text{and} \qq \gchar{edge\_connectivity}>1
\eqx
It is worth noting that the vast majority of the graphs obtained in the learning process have significantly larger \gchar{pca\_elongation} than the $0.25$ value appearing in the definition of Q1D (see bottom row in Table~\ref{tab.graphchar}). 
It is evident from the pictures in Fig.~\ref{fig.networks}, that the networks for CMNIST, CKMNIST and CFashionMNIST have clearly a quasi-1-dimensional structure.\footnote{Q1D is sometimes violated for the simplest networks which can be split by cutting a single edge.} One should emphasise that the initial network architecture -- the fully connected DAG -- has $\gchar{pca\_elongation}=0$, which means that the quasi-1-dimensional structure emerges in the process of learning.

\begin{wrapfigure}[13]{r}{0.31\textwidth}
\vspace{-0.8cm}
\centerline{\includegraphics[width=0.8\linewidth]{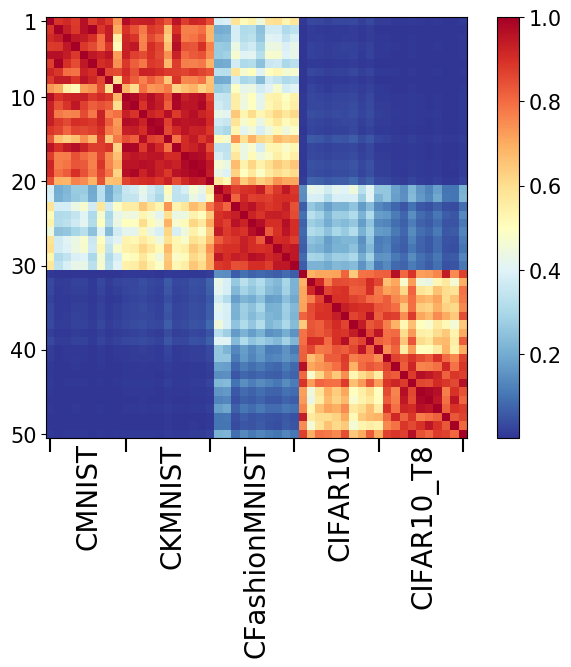}}
\caption{Network similarity matrix (see also Fig. 6 in {\it Supplementary Materials B}).}
\label{fig.similarity}
\end{wrapfigure}

In order to visualise to what extent the obtained networks for a given dataset are similar between themselves and different from networks for other datasets, for each of the 50 obtained networks we have collected the features exhibited in Tables~\ref{tab.structural} and \ref{tab.graphchar} (apart from the number of parameters), standardised them and evaluated pairwise similarity using the standard RBF kernel. The result is shown in Fig.~\ref{fig.similarity}. One may observe that for the same dataset, all the obtained graphs are strongly correlated. Moreover, networks working on tasks of comparable difficulty also seem to share similarities like CMNIST and CKMNIST, or CIFAR10 and CIFAR10\_T8. Yet in the latter case, one can nevertheless clearly distinguish the two groups of graphs for each dataset. 
This confirms that the uncovered wiring topology is firmly related to the problem it is solving.

\paragraph{Graph characterisation of datasets.}

Let us finally mention an intriguing, purely theoretical application of the techniques employed in the present paper. 
For each dataset one obtains a series of characteristic graph structures of neural networks which optimise the performance versus resources trade off.
Therefore we obtain a mapping
\eq
\DD \longrightarrow P(\GG)
\eqx
which associates to each dataset $\DD$ (more precisely learning task) a distribution of graphs. Moreover this probability distribution seems to be quite localised, as we have observed that the obtained networks have consistently characteristic features for a given dataset (see Fig.~\ref{fig.similarity}).
It is tempting to speculate, that the graphs represent in some way the natural hidden semantic structure of the given classification task.\footnote{This yields complementary neural network based characterisations of datasets to the ones discussed in~\cite{complexity}.} 
It would be very interesting to explore this further.

\section{Conclusions and outlook}

In the present paper we have implemented a conceptually simple but effective method of learning the network's size and topology simultaneously with learning the classification task. 
By basing the construction on an initial fully connected DAG, the network has full freedom of exploring any kind of local and global connectivity compatible with the feed-forward character of information processing.

The resulting networks are quite sparse, yet they do not require any subsequent fine-tuning retaining virtually identical test accuracy to the unthresholded network. Moreover, training the reduced networks from scratch with arbitrary initialisation also yields same performance.
This is in contrast to many pruning approaches, but it supports the hypothesis of~\cite{liu2018rethinking} that structural pruning can uncover optimal architectures.

The method proposed in the present paper yields networks whose size and complexity is clearly correlated with the difficulty of the dataset. On the one hand, we observe certain common features like the more involved connectivity of the high-level processing and a quasi-1-dimensional structure emerging for many of these networks. On the other hand, we observe differences specific to particular datasets like the proportion of nodes in various resolution stages etc.

Finally, the fact that the obtained graphs are characteristic to the specific learning tasks opens up a fascinating possibility of understanding them as representing hidden internal structures of the particular datasets.

\vfill

\pagebreak

\paragraph{Acknowledgements.} This work was supported by the Foundation for Polish Science (FNP) project \emph{Bio-inspired Artificial Neural Networks} POIR.04.04.00-00-14DE/18-00.



\bibliographystyle{unsrt}
\bibliography{main}

\section*{Supplementary Materials}

\appendix

\section{Training time}

In Table \ref{tab:epoch_time} we report the mean time of one epoch during the training of the fully connected DAG network and the retrained networks obtained for threshold $0.006$. In the latter case, as described in the paper, we consider two situations: retraining with the same number of initial channels $C=11$ as the original, complete DAG and fitting the number of channels $C$ separately for each architecture in order for the networks to have approximately the same number of \emph{parameters} as the complete DAG. All models were trained using the GeForce RTX 2080 Ti graphic card. Although for different computational infrastructure the values may vary a bit, the advantage gained after pruning is indisputable. 

\begin{table}[h]
    \centering
    \caption{The mean epoch time during the training of the original graph, the obtained network with number of initial channels $C$ being set to the same value as the fully connected DAG, and the obtained network when $C$ was fitted separately for each architecture. }
    \vskip 0.1in
\begin{tabular}{llll}
\toprule
{} & full DAG & rerun & rerun (fit C) \\
dataset        &               &                    &                     \\
\midrule
CMNIST        &  151.99$\pm$17.73 &         21.65$\pm$3.28 &          21.81$\pm$3.60 \\
CKMNIST        &  147.03$\pm$19.55 &         26.05$\pm$4.39 &          25.84$\pm$7.00 \\
CFashionMNIST  &  149.71$\pm$19.46 &        41.09$\pm$16.92 &          32.77$\pm$2.33 \\
CIFAR10        &  120.98$\pm$16.59 &         75.29$\pm$3.07 &         59.23$\pm$13.23 \\
CIFAR10\_T8 &   107.42$\pm$5.53 &         76.09$\pm$3.63 &          69.45$\pm$5.48 \\
\bottomrule
\end{tabular}
    \label{tab:epoch_time}
\end{table}

\section{Networks obtained for various datasets}

\begin{figure}[h]
    \centering
    \includegraphics[width=0.4\textwidth]{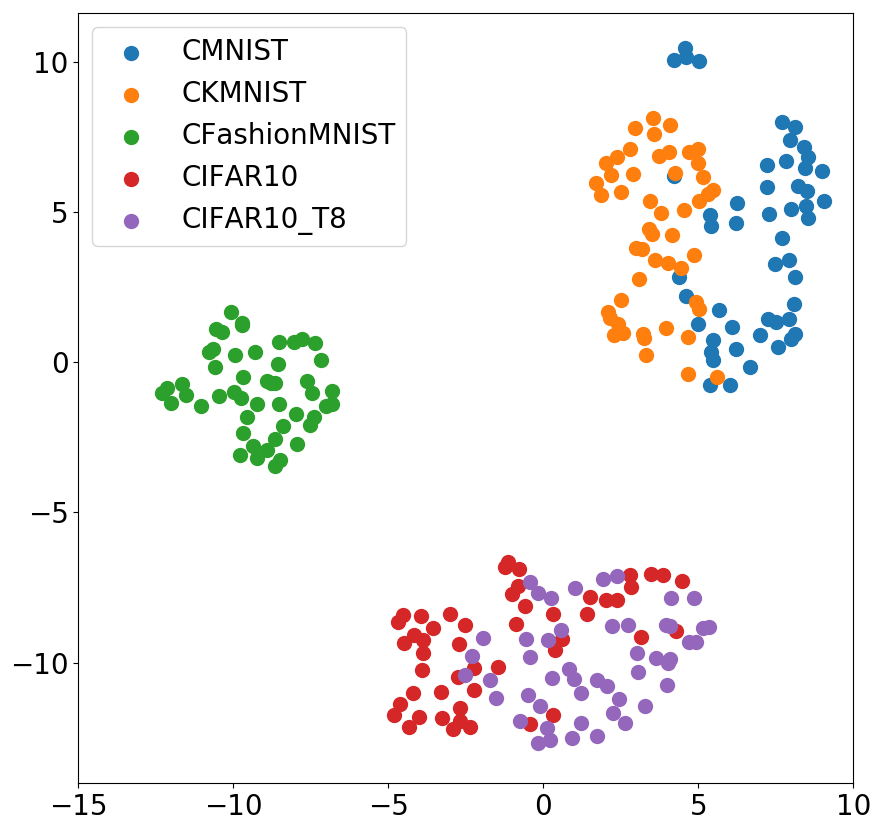}
    \caption{The UMAP embedding of the networks obtained for thresholds $\tau \le 0.006$. Different colours represent different datasets. }
    \label{fig:umap}
\end{figure}

\begin{figure}[thb!]
  \centering

\begin{tabular}{ccccc}
\rotatebox[origin=l]{90}{CMNIST} &
\includegraphics[height=3.75cm]{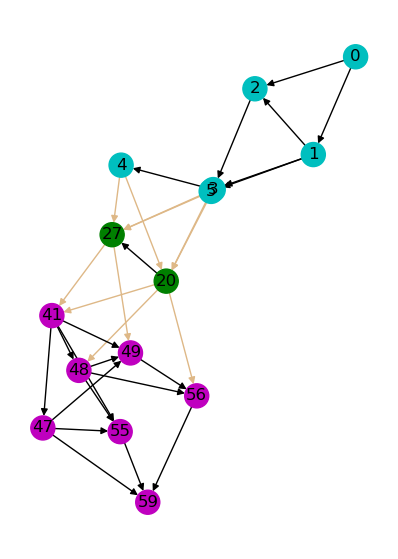} &
\includegraphics[height=3.75cm]{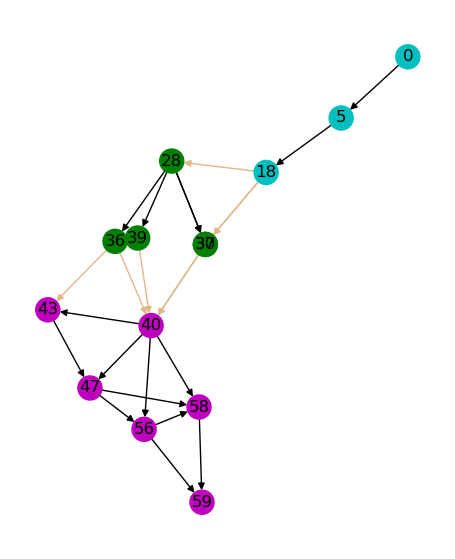} &
\includegraphics[height=3.75cm]{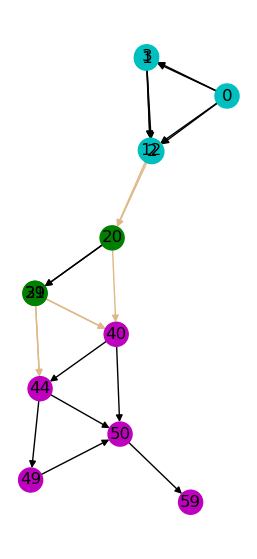} &
\includegraphics[height=3.75cm]{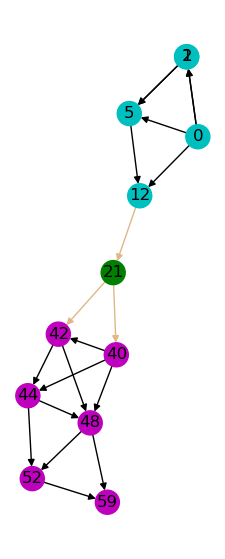} \\

\rotatebox[origin=l]{90}{CKMNIST} &
\includegraphics[height=3.75cm]{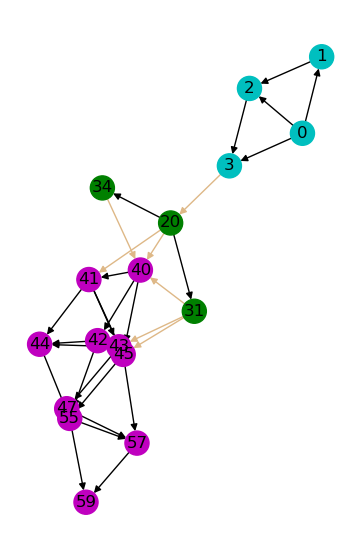} &
\includegraphics[height=3.75cm]{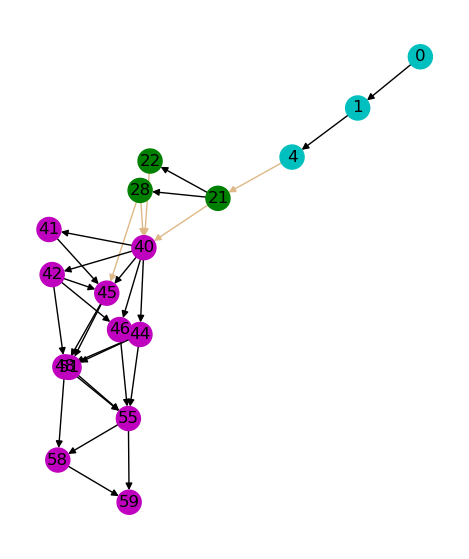} &
\includegraphics[height=3.75cm]{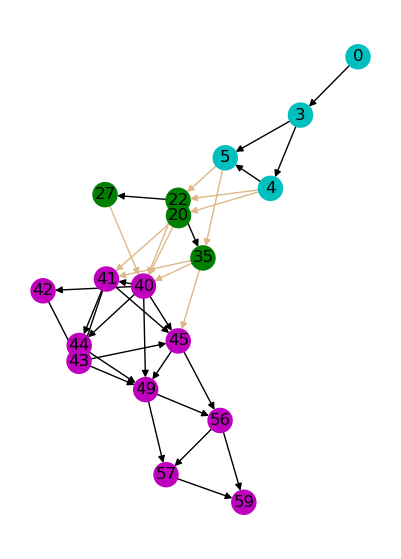} &
\includegraphics[height=3.75cm]{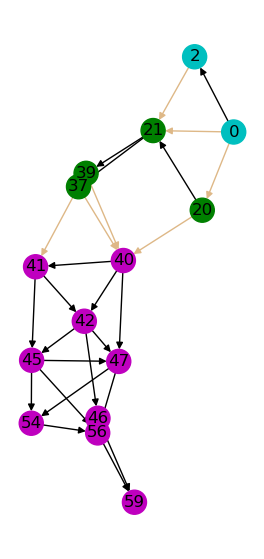} \\

\rotatebox[origin=l]{90}{CFashionMNIST} &
\includegraphics[height=3.75cm]{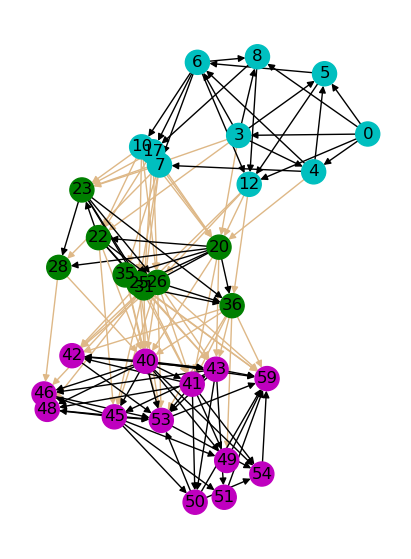} &
\includegraphics[height=3.75cm]{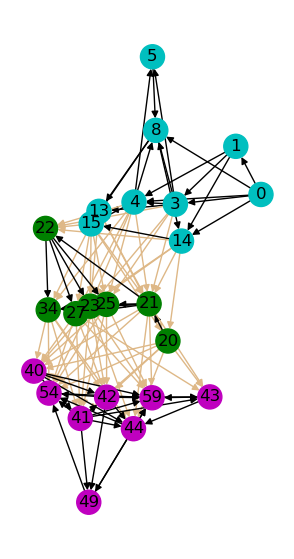} &
\includegraphics[height=3.75cm]{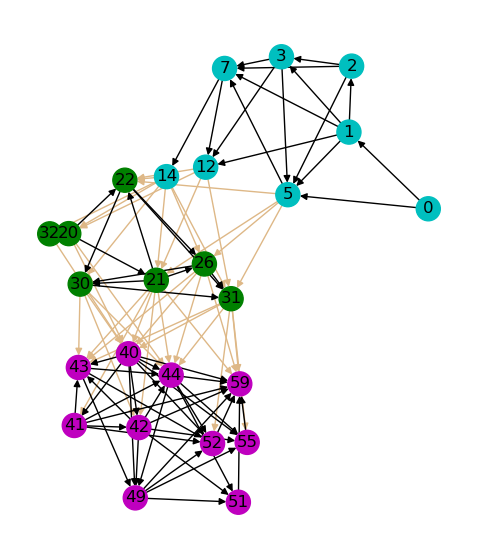} &
\includegraphics[height=3.75cm]{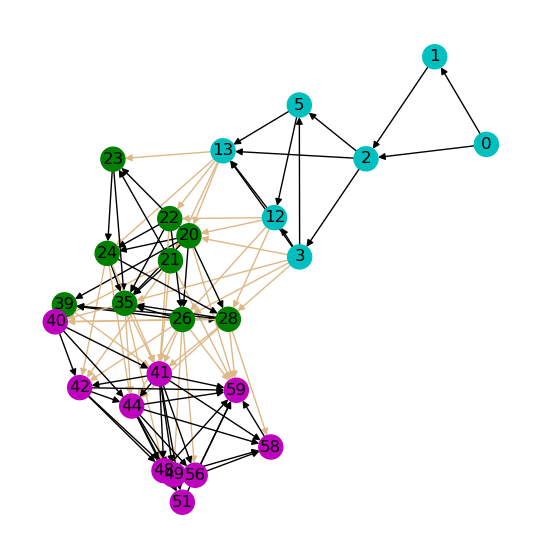} \\

\rotatebox[origin=l]{90}{CIFAR10} &
\includegraphics[height=3.75cm]{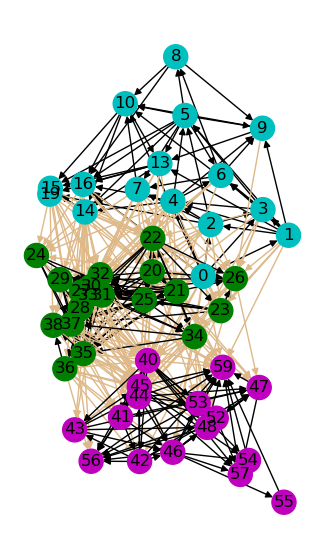} &
\includegraphics[height=3.75cm]{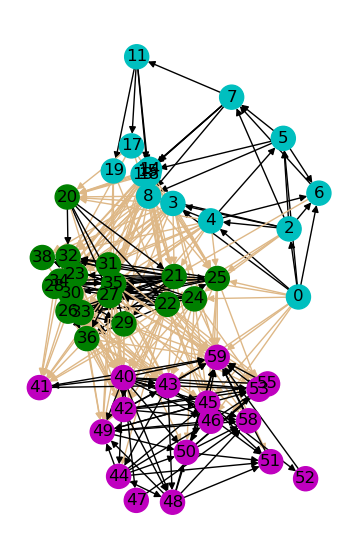} &
\includegraphics[height=3.75cm]{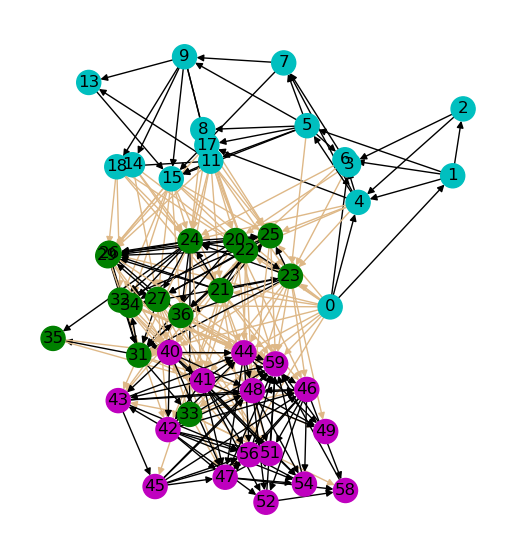} &
\includegraphics[height=3.75cm]{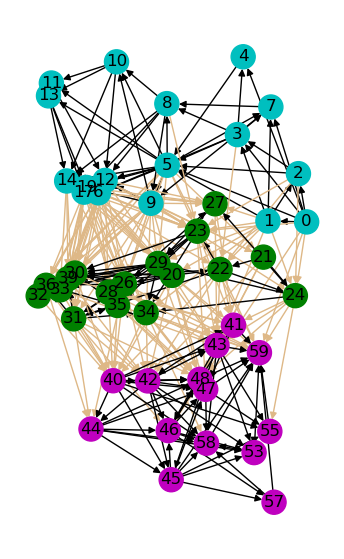} \\

\rotatebox[origin=l]{90}{CIFAR10\_T8} &
\includegraphics[height=3.75cm]{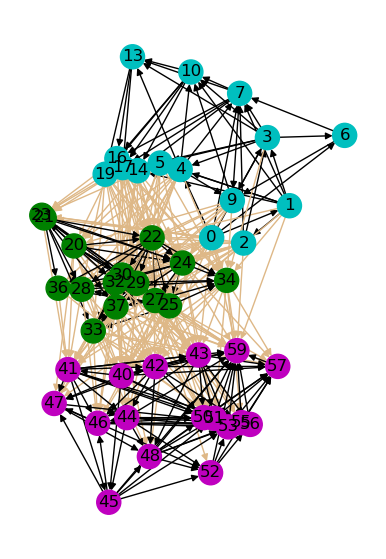} &
\includegraphics[height=3.75cm]{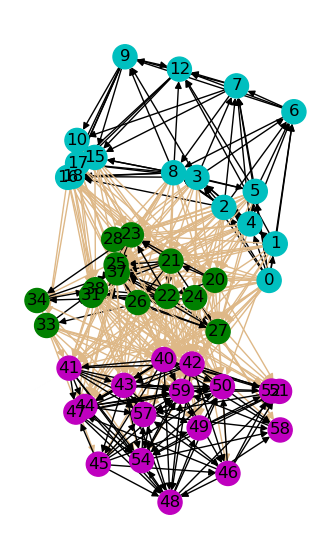} &
\includegraphics[height=3.75cm]{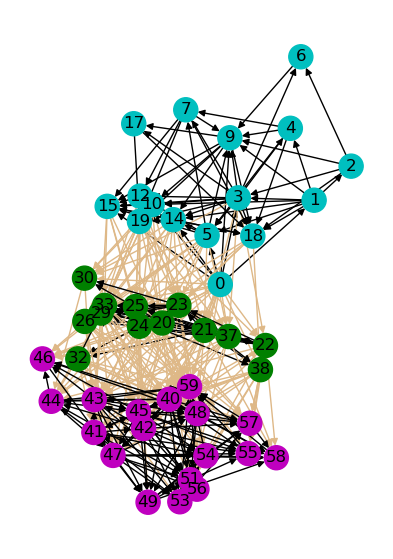} &
\includegraphics[height=3.75cm]{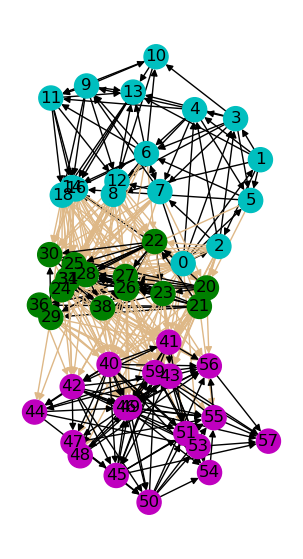} \\

\end{tabular}
 
  \caption{The thresholded networks (at the threshold 0.006) obtained from training with several random initialisations.}
  \label{fig.networkspage}
\end{figure}

In Fig. \ref{fig:umap} we present the UMAP embedding of the obtained architectures for thresholds equal or smaller than $0.006$. The embedding was computed on the standardised features from Tabels 1 and 2 from the paper (apart from the number of parameters). It is evident that networks solving the same task tend to cluster, being similar to each other. This similarity can be also visually observed in Fig.~\ref{fig.networkspage}, where we demonstrate a subset of the obtained networks for threshold $0.006$ for different datasets with various initialisations.

\section{Training setup}

We train the fully connected DAG architectures 10 times for each dataset using different initialisation but the same learning regime. All models are trained for 100 epochs using the SGD algorithm with starting learning rate 0.1, momentum 0.9, batch size 128 and weight decay 1e-4. In the 80th and 90th epoch the learning rate is decreased by factor 10. For the retrain experiments we use the same settings but with different initialisation seed. All networks were trained using the GeForce RTX 2080~Ti graphic card.

\vfill
\pagebreak

\end{document}